\documentclass[10pt,twocolumn,letterpaper]{article}

\usepackage{cvpr}              

\usepackage{graphicx}
\usepackage{amsmath}
\usepackage{amssymb}
\usepackage{booktabs}

\usepackage[pagebackref,breaklinks,colorlinks]{hyperref}

\usepackage[capitalize]{cleveref}
\crefname{section}{Sec.}{Secs.}
\Crefname{section}{Section}{Sections}
\Crefname{table}{Table}{Tables}
\crefname{table}{Tab.}{Tabs.}

\def\cvprPaperID{7859} 
\def\confName{CVPR}
\def\confYear{2023}

\begin{document}

\title{Evaluation of autonomous systems under data distribution shifts}

\author{Daniel Sikar\\
City, University of London\\
Northampton Square, London EC1V 0HB, UK\\
{\tt\small daniel.sikar@city.ac.uk}
\and
Artur Garcez\\
City, University of London\\
Northampton Square, London EC1V 0HB, UK\\
{\tt\small A.GARCEZ@city.ac.uk}
}
\maketitle


\begin{abstract}
We posit that data can only be safe to use up to a certain threshold of the data distribution shift, after which control must be relinquished by the autonomous system and operation halted or handed to a human operator.
With the use of a computer vision toy example we demonstrate that network predictive accuracy is impacted by data distribution shifts and propose distance metrics between training and testing data to define safe operation limits within said shifts.
We conclude that beyond an empirically obtained threshold of the data distribution shift, it is unreasonable to expect network predictive accuracy not to degrade.

\end{abstract}


\section{Introduction}

The development of autonomous systems such as self-driving cars is motivated by a number of goals, of a practical, safety, public interest and economic nature.
From a practical perspective, the goal is "to transport people from one place to another without any help from a driver" \cite{s20092544}.
From a public health perspective, to transform the current approach
to automotive safety from reducing injuries after collisions to
 complete collision prevention \cite{Fleetwood_2017}.
From a public interest and economic perspective, AV fleets allow for new shared autonomous mobility business models 
\cite{riggs2019business}
though shared autonomous electric vehicle (SAEVs) fleets \cite{loeb2019fleet}. Shared Autonomous Vehicles (SAVs) have gained significant public interest as a possible less expensive, safer and more efficient version of today’s transportation networking companies (TNCs) and taxis.

This perceived superiority to human drivers is attributed to high-performance computing that allows AVs to process, learn from and adjust their guidance systems according
to changes in external conditions at much faster rates than the typical human driver  \cite{west2016moving}.

The presence in public spaces of autonomous systems  driven by AI is a concern. Although models such as convolutional neural networks have been successfully used to solve problems applied to Computer Vision, the ability to generalize and robustness of such architectures has been increasingly scrutinised. 

Zhang \etal \cite{zhang2017understanding} debated the need to rethink generalization, by demonstrating how traditional benchmarking approaches fail to explain why large neural networks generalize well in practice. By randomizing target labels, the experiments show that state-of-the-art convolutional neural networks for image classification trained with SGD (stochastic gradient descent) are large enough to fit a random labelling of the training data. This is achieved with a simple two-layer neural network, which presents a "perfect finite sample expressivity" once the number of parameters is greater than the number of data points as often is the case with CNNs.  
This poses a challenge, in the context of overfitting, for autonomous systems such as self-driving cars, specially if relying only on images for perception of the surrounding environment. 
Since testing real life scenarios is not practical due to the associated cost and risk, we examine the use of game engines \cite{cowan2014survey}, which are capable of generating labeled datasets and realistic environments where the autonomous vehicle may be tested.



In this study, data distribution shifts are examined in the context of testing data presented to a neural network trained on a known data distribution, that is, with the weights and biases adjusted while minimizing an error function. 

\begin{figure*}
\centering
\includegraphics[width=\textwidth]{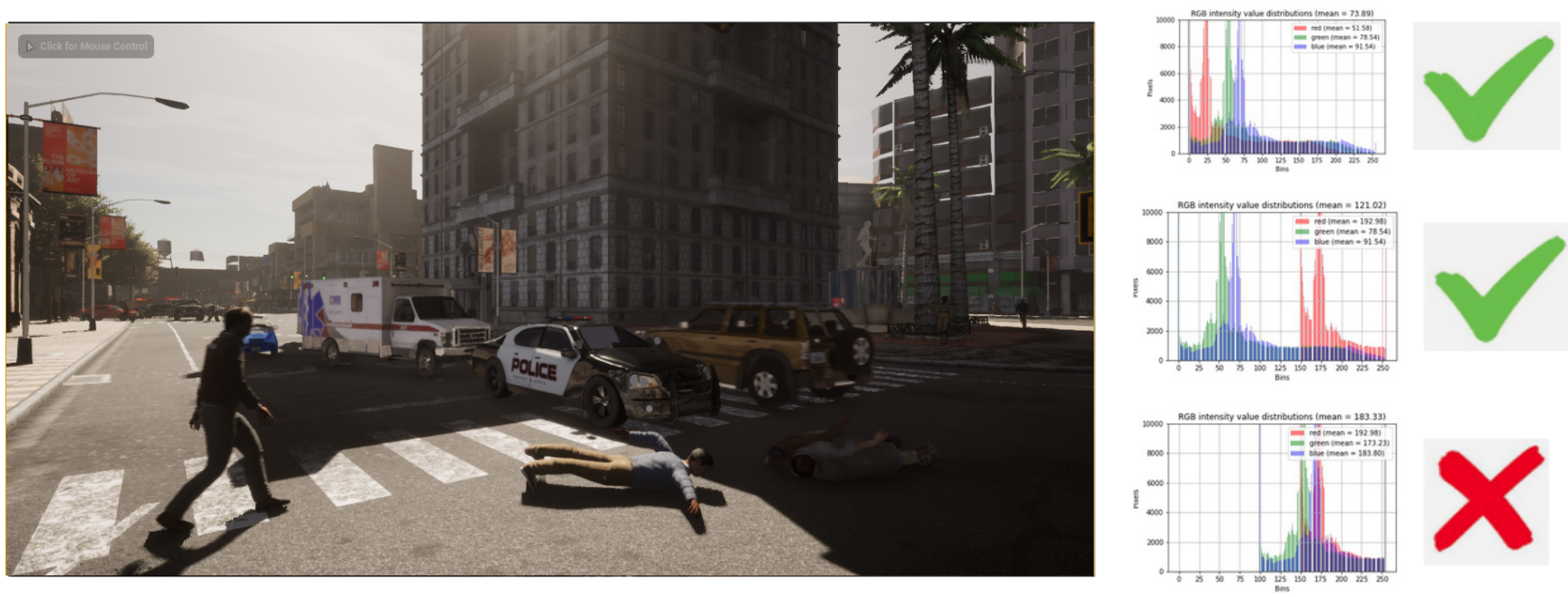}
\caption{Simulated accident in the CARLA Simulator Town 10, where excessive brightness i.e. high RGB values cause predictive accuracy of self-driving models to degrade}
\label{fig:CarlaTicks}
\end{figure*}

When the shift is of a sufficient quantity, relative to the training data, a number of terms exist in the literature to express and deal with the shift, such as anomaly detection \cite{chalapathy2019deep,bulusu2020anomalous,pang2021deep,ruff2021unifying}, out-of-distribution data \cite{fort2021exploring, shalev2018out, huang2021mos, hsu2020generalized, hendrycks2016baseline}, novelty detection \cite{tack2020csi,pidhorskyi2018generative,kerner2019novelty,diehl2002real,pimentel2014review}, outlier detection \cite{basu2007automatic,hodge2004survey,aggarwal2001outlier}
and covariate shift \cite{sugiyama2007direct,sugiyama2012machine,quinonero2008dataset, bickel2009discriminative}. In section \ref{sec:Methods} we propose metrics to quantify the distance between training and testing data. We use a game engine to generate our training data and train a model based on SOTA self-driving architectures. Our aim is to determine safe i.e. acceptable shift quantities, where predictions made by the model, given the shifted data, are safe to use, as exemplified in Figure  \ref{fig:CarlaTicks}.

\section{Related Work}

We examined literature related to out-of-distribution (OOD) data, and methods for identifying such data and/or making neural networks robust to its adverse and unwanted effect, such that network predictive accuracy is not affected by data in the OOD regime.

Fort \etal \cite{fort2021exploring}
are motivated by out-of-distribution detection (OOD) and point out that is high-stake applications such as healthcare and self-driving. Techniques for detecting OOD inputs using neural networks include MSP (maximum over softmax probabilities) proposed by Hendrycks \etal \cite{hendrycks2016baseline} i.e. $score_{msp}(x)=max_{c=1,...,K}p(y=c|x)$, showing that the probability prediction for correct examples tends to be higher than of incorrect and out-of-distribution examples. Capturing prediction probability statistics about in-sample examples is in most cases sufficient for detecting abnormal examples.

Lee \etal \cite{lee2018simple} argue that Mahalanobis distance is more effective than Euclidean distance in various OOD and adversarial data detection tasks, given a defined distance-based confidence score $M(x)$ that is the Mahalanobis distance between a test sample \textbf{x} and the closest class-conditional Gaussian distribution:

\begin{equation*}
M(x) = \max\limits_c - (f(x) - \hat{\mu}_c)^T\hat{\sum}^{-1}(f(x)-\hat{\mu}_c).
\end{equation*}

\begin{figure*}
\centering
\includegraphics[width=\textwidth]{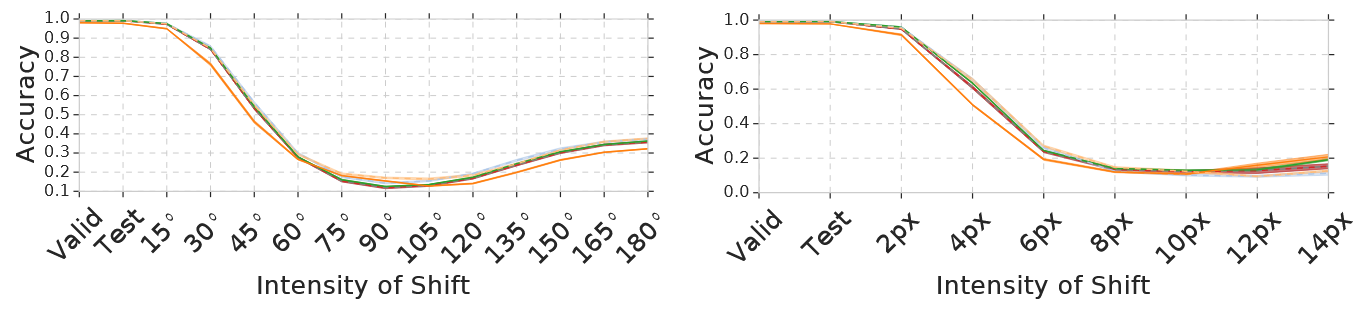}
\caption{LeNet predictive accuracy change under rotation (left) and translation shift (right)}
\label{fig:Ovadia2019}
\end{figure*}

Noting that the metric corresponds to measuring the log of the probability densities of the test sample. The value $\hat{\mu}_c$ represents the empirical class mean while $\hat{\sum}$ the class covariance of the training samples. The algorithm uses the weights and biases of the softmax classifier, that is the penultimate (fully connected, dense) layer and assumes that the class-conditional distribution follows a multivariate Gaussian distribution. 

Shalev \etal \cite{shalev2018out} propose the use of a number of distinct word embeddings as supervisors in ensemble models, thus generating shared representations. A semantic structure is utilised for word embeddings to produce semantic predictions, while the L2-norm of the output vectors is employed to detect OOD inputs, testing the approach to detect adversarial and incorrectly classified examples. \textit{Cosine distance} is used in the space $Z$ to measure the distance between two embeddings \textbf{u},\textbf{v} $\in Z$:

\begin{equation*}
d_{cos}\mathbf{(u,v)=\frac{1}{2}\bigg(1-\frac{u\cdot v}{\lVert u \lVert \lVert v \lVert}\bigg)}
\end{equation*}

When the cosine distance is close to 0, the labels are considered similar, when it is close to 1, the labels are considered to be semantically far apart.

Huang \etal \cite{huang2021mos} propose a scoring function based on group softmax for different categories with the \textit{Minimum Others Score} (\textbf{MOS}) function, allowing differentiation between in and out-of-distribution data. The key observation being that a pre-defined category \textit{others} carries useful information for the likelihood of a given imaged being OOD w.r.t. each group. An OOD input is mapped to \textit{others} in all groups and the lowest \textit{others} score in all groups determines OOD images. The \textbf{MOS} OOD scoring function being: 

\begin{equation*}
S_{MOS}(\mathbf{x}) = - \min\limits_{1\leq k \leq K} p_{others}^k(\mathbf{x})
\end{equation*}

The sign is negated such that $S_{MOS}(\mathbf{x}) $ is higher for in-distribution and lower of out-of-distribution data.

Liang \etal \cite{liang2017enhancing} propose the Out-of-distribution detector (ODIN) to distinguish in and out of distribution images on a pre-trained neural network using a temperature scaling parameter $T$. The problem statement is defined by $P_{\mathbf{x}}$ and $Q_{\mathbf{x}}$ denoting two distinct data distributions defined on the image space $\chi$, where $P_{\mathbf{x}}$ are the in-distribution and $Q_{\mathbf{x}}$ are the out-of-distribution images. During training images are drawn from $P_{\mathbf{x}}$. If testing images are drawn from a mixture distribution, can a distinction be made between in an out-of-distribution images?  
For each input the network generates a label prediction  $\hat{y}(x)= \arg \max_i S_i(x;T)$ by computing the softmax output for each class, using the temperature scaling term $T$ set to 1 during training:

\begin{equation*}
S_i(x;T) = \frac{\exp (f_i(x) / T)}{\sum_{j=1}^N \exp (f_j(x) / T) }
\end{equation*}

Temperature scaling perhaps resonates with von Neumann \cite{von1956probabilistic} whose conviction was that error should be treated and subject to themodynamical methods and theory. In addition to temperature scaling, a small perturbation $\epsilon $, inspired by the idea of adversarial examples \cite{goodfellow2014explaining}, is added to the inputs. The inputs are classified as in-distribution if the softmax score is greater than threshold $ \delta $, out-of-distribution otherwise. The out-of-distribution can therefore be defined as a function of the input $x$, the perturbation $\epsilon$, the temperature scaling term $ T$ and the threshold $\delta$:

\begin{equation*}
  g(x;\delta,T,\epsilon) =
    \begin{cases}
      1 & \text{if $\max_i p(\bar{x};T) \leq \delta $  },\\
      0 & \text{if $\max_i p(\bar{x};T) > \delta $  }.
    \end{cases}       
\end{equation*}

Hsu \etal \cite{hsu2020generalized} use a combination of ODIN and Mahalanobis distance for OOD data detection. Pointing out that ODIN requires OOD data to tune hyperparameters for temperature scaling and input pre-processing, their Generalized ODIN method addresses the problem of learning without OOD data, adding a variable to declaring if the data is in or out of distribution, closely related to the temperature scaling term in ODIN, except that the variable depends on input instead of tuned hyperparameters.



Ovadia \etal \cite{ovadia2019can} argue that high stake applications such as medical diagnoses and self-driving require in addition to predictive accuracy, quantification of predictive uncertainty, that is, class predictions and confidence values. The study concluded that accuracy and confidence degrade with dataset shifts (rotated or horizontally translated images), and that ensembles help with robustness to dataset shifts. Figure \ref{fig:Ovadia2019} shows accuracy shift as a function of rotated and translated MNIST datasets, as predicted with the LeNet architecture  \cite{lecun1998gradient} with standard training,
validation, testing and hyperparameter tuning protocols, and seven variants including vanilla, ensemble and dropout. The work presented by Ovadia \etal \cite{ovadia2019can} with respect to accuracy and dataset shifts is the one we find our study most aligned with.



\section{Methods}
\label{sec:Methods}

\begin{figure*}
\centering
\includegraphics[width=\textwidth]{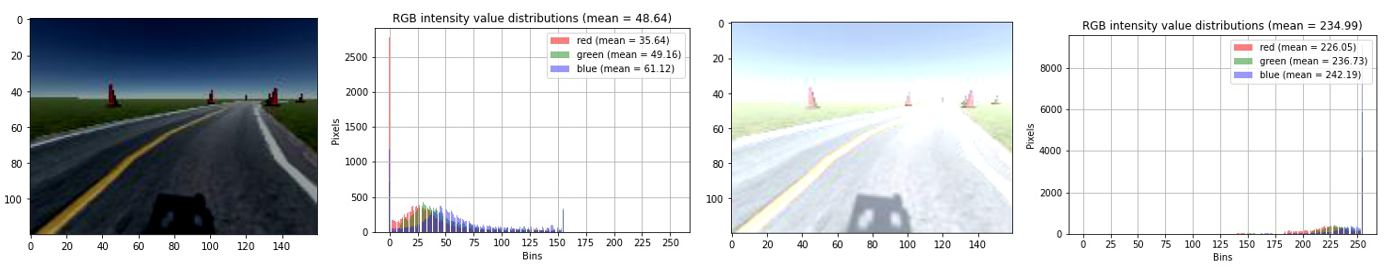}
\caption{Two sets of images and pixel intensity value histograms, where the set on the left is a negative shift (pixel intensity values decrease) and the set on the right is a positive shift (pixel intensity values increase)}
\label{fig:RGBShiftsMinusAndPlus}
\end{figure*}

\subsection{Applying data distribution shifts to RGB images}
\label{methods:applying_data_distribution_shifts}

We interpret an RGB image as a data distribution of individual pixel intensity values. This can be represented as a histogram where the sum of bucket counts (distinct pixel values aggregated counts) is given in Equation \ref{eq:t_p}, where the number of buckets is equal to the number of values the pixel may take, i.e. the pixel datatype range of values.
Distribution shifts therefore will cause counts to increase and decrease in each bucket, according to the direction (left, negative, right, positive) of distribution shift.
The distribution is shifted by uniformly incrementing or decrementing each pixel intensity value by the same quantity. We define the RGB pixel intensity shift function as:

\begin{equation}
\label{eq:pixel_intensity_shift}
    RGB_{is}(I,S) = 
    \begin{cases}
     I_{jk} + S, & \text{if $I_{imin} \leq I_{jk} + S \leq I_{imax}$ }.\\ 
    I_{imin}, & \text{if $I_{imin} > I_{jk} + S $}.\\
    I_{imax}, & \text{if $I_{max} < I_{jk} + S $}.
    \end{cases}
\end{equation}

where $I$ is the RGB image pixel matrix, $S$ is the shift quantity to be added to each pixel, $j$ is the number of dimensions in $ I$ and $k$ is the number of elements in $j$. $S$ takes both negative and positive values. Negative values cause a left shift to the distribution and decrease the pixel intensity value. Output values can be in the range of, and including, $I_{imin}, I_{imax}$ defined respectively as the minimum and maximum values the pixel data type can represent.
A 24-bit RGB image, represented by 3 bytes, one for every red, green and blue channel, can store values between 0 and 255, in this case, if the value of $I_{jk}$ is 100 and a shift of -120 is applied, $I_{jk}$ will take value 0. If $I_{jk}$ is 170 and a shift of 100 is applied $I_{jk}$ will take value 255. We say in these cases that the pixel is saturated resulting and darker or brighter images. The result and effect can be observed in Figure \ref{fig:RGBShiftsMinusAndPlus}.

\begin{figure*}
\centering
\includegraphics[width=\textwidth]{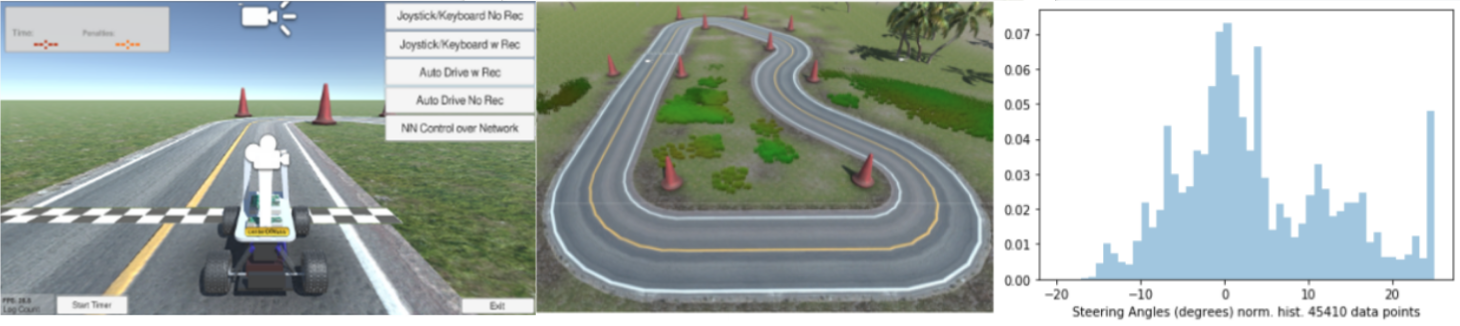}
\caption{Left to right, the SDSandbox self-driving neural network training application, the Generated Track circuit, a steering angle histogram showing the distribution of steering angles when going around the track clockwise}
\label{fig:SDSandbox_track_distribution}
\end{figure*}

\subsection{Image pre-processing}

One of the key pre-processing steps used in SOTA models such as \cite{lecun2004dave} is moving images from RGB to YUV space. The authors do not discuss the motivation behind this procedure. [YUV article] points out that YUV is more related to human vision. It can be said that YUV reduces the information content of the image. We observed that when no datashift exists, the YUV mean is lower than the RGB mean, and we decrease the RGB mean. we observed that after that when with a mean shift of minus 5 the RGB and YUV means are equal. Increased negative shift leads to an increase in the distance between RGB and YUV means.

We start with an image. We take the mean value of all combined RGB channel values, we then move the image to YUV space and back to RGB space, then again compute the mean. When there is no shift in pixel intensity value, the YUV image has a lower mean than the corresponding RGB mean. Depending on the shift, and the image becoming darker, or brighter, the YUV mean may be higher (for darker images, or lower (for brighter images). The main takeaway being the mean delta is lower in YUV space, creating lesser variability in the data and validating the use of YUV images to train self-driving CNNs and potentially other applications and architectures. This finding is a side-effect, and not the main focus of this study.

One important aspect of training data for AVs is data pre-processing e.g. cropping such that most relevant section of the image to the self-driving task is kept. In the example shown, 25 rows of pixels were removed from the bottom and 60 rows of pixels were removed from the top, resulting in a 75h x 320w pixel image. The image is then resized to 66h x 200w, the geometry to be presented to network in the example shown. Finally the image is moved from RGB to YUV space. In the RGB scheme, each pixel is represented by three channel intensities of red, green and blue. In YUV, also referred to as YCbCr space, each pixel is represented by Y (luma), U (Cb - luminance value subtracted from blue channel) and V (Cr - luminance value subtracted from red channel). It is a "lossy" process which degrades the data, and originally developed for colour to black-and-white television backward-compatibility. Moving the image from RGB to YUV space has been demonstrated to give "better subjective image quality than the RGB color space", being better for computer vision "implementations than RGB due to the perceptual similarities to the human vision" \cite{podpora2014yuv}. This scheme was used in previous AV data pre-processing pipelines e.g. Dave \cite{lecun2004dave} and PilotNet \cite{bojarski2016end}.

\subsection{Game engines}
\label{methods:GameEngines}

Game engines are ubiquitous and the concept of using virtual environment in lieu of real ones, to train autonomous systems or humans, with the benefits of cost savings and security, is long established. We used a Unity game engine based self driving training and evaluation system called SDSandbox, short for self-driving sandbox.
Figure \ref{fig:SDSandbox_track_distribution}  shows left to right the SDSandbox user interface, with options including to drive, self-drive with a PID algorithm and self-drive with a neural network. In the middle is the Generated Track and on the left a histogram of steering angles for a dataset generated by the SDSandbox driving around the generated track, where the autonomous vehicle drives clockwise around the track, and the start is on the top left of the circuit.

The Unity SDSandbox can generate a number of circuits. We obtained a labelled dataset of 20,061 frames for the Generated Road circuit and a labelled dataset of 1,394 frames for the Generated Track circuit, each frame being a jpg image of size 160 wide by 120 high, with a corresponding json file containing the steering angle and throttle applied at the time the image was saved.  
We trained networks to self-drive in the SDSandbox environment, a video of two Donkey Cars \cite{DonkeyCar2022} being driven by distinct trained networks, on a similar random Generated Track, can be seen at \cite{Donkeyx2}. Note the video is different in that the frames are generated by the video game in real time, and the simulated world adjusted accordingly, depending on the steering and position of simulated vehicle on the road. With the labelled datasets, the world is not adjusted, as the frames have been generated and saved previously, and whatever steering prediction the model generates, it will not actually change the next frame, as it would in realtime.

\begin{figure*}
\centering
\includegraphics[width=\textwidth]{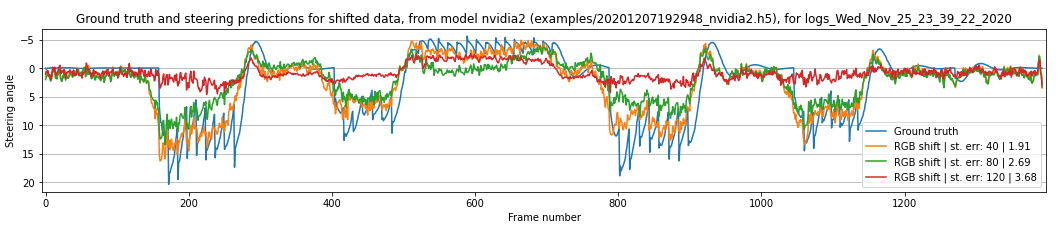}
\caption{Plots of ground truth (SDSandbox PID steering output) and nvidia2 network predictions for images with pixel value intensity shifts of 40, 80 and 120, where the steering error (st. err.) for RGB shift is the MAE }
\label{fig:dataShiftedSteeringPredictions-p40p80p120}
\end{figure*}

\subsection{Metrics}
\label{methods:metrics}

We examine 4 metrics for defining network prediction error, and 3 metrics for defining the distance between training and testing data distributions.

\subsubsection{Error functions}
\label{methods:error_functions}
\begin{equation}
    MAE = \frac{1}{N}\sum\limits_{i=1}^{N}|y_i-\hat{y}_i|  
\end{equation}

\begin{equation}
    MAPE = \frac{1}{N}\sum\limits_{i=1}^{N}\big|\frac{y_i-\hat{y}_i}{y_i}\big|
\end{equation}

\begin{equation}
    RMSE = \sqrt{\frac{1}{N}\sum\limits_{i=1}^{N}(y_i-\hat{y}_i)^2}
\end{equation}

\begin{equation}
    MSE = \frac{1}{N}\sum\limits_{i=1}^{N}(y_i-\hat{y}_i)^2
\end{equation}

Where MAE is the Mean Absolute Error, MAPE is Mean Absolute Percentage Error, MSE is Mean Squared Error and RMSE is Root Mean Square Error. Where MSE is quadratic, i.e., penalising larger errors while the other three are linear. $y$ is the ground truth while $\hat{y}$ is the network prediction value, and $N$ is the total number of images presented to the network, over which the error is computed




\subsubsection{Distance functions}
\label{methods:distance_functions}

We adapt three known distance functions to quantify distance between two images, where the first is assumed to have been used during training, that is of a known accuracy and not have expected to generate a steering value that would have caused the simulated vehicle to drive off the road.

We consider an RGB image as distributions, as represented in a histogram, corresponding to each red, green and blue channels where the ground truth and shifted RGB histogram bin counts are $P,Q$ respectively.

We use the product $t_p$ obtained in equation \ref{eq:t_p} to normalise our bin counts $ P, Q $ where $ N $ is the number of dimensions in the image array $ I $ and $ I_n $ is the number of elements in each image dimension $n$.


\begin{equation}
\label{eq:t_p}
  t_p = \prod\limits_{n=1}^{N}I_n   
\end{equation}


\textbf{Bhattacharyya RGB distance} we define as the negative natural logarithm of the Batthacharyya RGB coefficient:

\begin{equation}
    D_Brgb(P,Q) = -ln ( BCrgb(P,Q)    
\end{equation}
where the Batthacharyya coefficient is defined as:

\begin{equation}
    BC(P,Q) = \sum\limits_{x\in \mathcal{X}} \sqrt{P(x) Q(x)}
\end{equation}

for every $x$ on the same domain $\mathcal{X}$. Since we are dealing with multiple dimensions (channels), in our Batthacharyya RGB coefficient we explicitly divide our histogram bin counts by $t_p$, and the expression becomes:

\begin{equation}
\begin{split}
BCrgb(P,Q) & = \sum\limits_{x\in \mathcal{X}} \sqrt{\frac{P(x)}{t_p} \frac{Q(x)}{t_p}} \\
& = \sum\limits_{x\in \mathcal{X}} \sqrt{\frac{P(x)Q(x)}{t_p^2} } \\
& = \sum\limits_{x\in \mathcal{X}} \frac{\sqrt{P(x)Q(x)}}{t_p}  \\
& = \sum\limits_{x\in \mathcal{X}} \sqrt{P(x)Q(x)} \;  t_p^{-1} 
\end{split}
\end{equation}

\textbf{Relative entropy}, also known as Kullback-Leibler divergence, for our specific use with RGB images we express as:

\begin{equation}
\begin{split}
 D_\text{KLrgb}(P \parallel Q) & = \sum\limits_{x\in\mathcal{X}} P(x)t_p^{-1} \log\left(\frac{(P(x)+\epsilon) t_p^{-1}}{(Q(x)+\epsilon)t_p^{-1}}\right)  \\
 & = \sum\limits_{x\in\mathcal{X}} P(x)t_p^{-1} \log\left(\frac{P(x)+\epsilon}{Q(x)+\epsilon}\right) 
\end{split}    
\end{equation}

We add a small positive term $ \epsilon $ to $ P(x), Q(x) $ to avoid division by zero, and also to avoid taking the logarithm base 10 of zero which is undefined.

We note that the KL Divergence in addition to, for our purposes, being fragile, is also asymmetric. that is the divergence from P to Q is not the same as the divergence from Q to P.

Note we use discrete rather than the continuous values. That is, we use the original byte value for each channel and the discrete relative entropy equation. If we convert the distribution to a PDF, that is, with all aggregated counts summing to one, we would use the continuous equation. The PDF seems like the better way to go as it is a normalised value, though if we are dealing with a known network, the input is always of the same the same dimensions and can be considered normative, that is the RGB mean is constrained by, and bound to, the input size.
\textbf{Histogram RGB intersection} between two distributions we define as: 

\begin{equation}
  H_{irgb}(P, Q) = t_p^{-1} \sum\limits_{x\in\mathcal{X}}argmin(P(x), Q(x))   
\end{equation}

If $P,Q$ are the same for every $ x $, then the summation will be equal to the product $t_p$, resulting in 1, meaning there is a complete overlap between the two distributions. If $argmin(P(x), Q(x))$ is equal to zero for every $x$, then the summation will be equal to zero meaning there is no overlap between the histograms.

Since the histogram RGB intersection does not take square roots or logs, it is the most computationally efficient of the three distance metrics we investigate.

\section{Experiments}

A total of 10,360,826 predictions were made for our datashift and accuracy analysis. 
Using both datasets, performing a total of 241 laps on each of the Generated Track and Generated Road circuits, to obtain MAE, MAPE and MSE for all steering predictions, similar to shown in Figure \ref{fig:dataShiftedSteeringPredictions-p40p80p120}, where a shift of positive 120 (red plot) clearly is not steering, and positive 80 is on the limit of following the ground truth PID steering angle. The plot shows that the limit of safety in this case is between positive 40 and positive 80, with approximately the same for negative shifts, which are not shown here due to space constraints.

\subsection{Distribution shift metrics}

We started our analysis by choosing one random image (1105\_cam-image\_array\_.jpg) from the Generated Track dataset. Using the equations described in section \ref{methods:distance_functions}, we computed the distance from the image to itself, which is one for the Histogram Intersection (complete overlap), and zero (distance) for Relative Entropy and Bhatthacharrya Distance. We then apply positive and negative shifts, for a total of 241 shifts, from negative 120 to positive 120.

\begin{figure*}
\centering
\includegraphics[width=\textwidth]{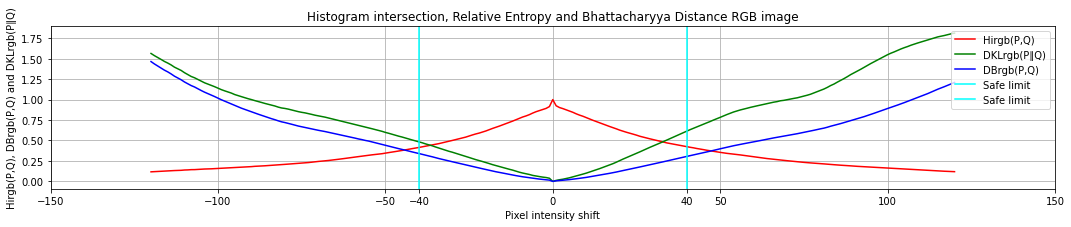}
\caption{Histogram Intersection, Relative Entropy and Bhattacharyya Distance safe limits for RGB image pixel intensity value shifts}
\label{fig:HI_KL_DB_safe}
\end{figure*}

\begin{table}

\caption{Histogram Intersection in RGB space between random Generated Road frames}
\label{table:hi_rgb_space}
\begin{center}
\resizebox{\columnwidth}{!}{%
\begin{tabular}{llllllllll}
\toprule
& ID1 & ID2 & -120 & -80 & -40 & 0 & 40 & 80 & 120 \\
\midrule
1 & 592 & 503 & 0.10 & 0.20 & 0.45 & 0.83 & 0.63 & 0.28 & 0.11 \\
2 & 863 & 825 & 0.08 & 0.21 & 0.55 & 0.80 & 0.37 & 0.16 & 0.07 \\
3 & 912 & 879 & 0.09 & 0.18 & 0.43 & 0.88 & 0.35 & 0.16 & 0.08 \\
4 & 1096 & 410 & 0.14 & 0.33 & 0.69 & 0.62 & 0.29 & 0.16 & 0.08 \\
5 & 519 & 772 & 0.11 & 0.29 & 0.62 & 0.81 & 0.44 & 0.18 & 0.09 \\
6 & 365 & 1091 & 0.10 & 0.19 & 0.32 & 0.65 & 0.68 & 0.35 & 0.14 \\
7 & 1082 & 1094 & 0.11 & 0.20 & 0.41 & 0.92 & 0.42 & 0.22 & 0.12 \\
8 & 811 & 1300 & 0.07 & 0.15 & 0.26 & 0.63 & 0.72 & 0.32 & 0.17 \\
9 & 146 & 1004 & 0.10 & 0.19 & 0.40 & 0.88 & 0.44 & 0.21 & 0.12 \\
10 & 1199 & 593 & 0.15 & \textbf{0.34} & 0.70 & 0.65 & 0.33 & 0.19 & 0.10 \\
11 & 797 & 1157 & 0.07 & 0.15 & \textbf{0.27} & \textbf{0.61} & \textbf{0.71} & 0.33 & 0.14 \\
12 & 350 & 173 & 0.08 & 0.16 & 0.29 & 0.65 & 0.72 & 0.36 & 0.15 \\
13 & 407 & 1260 & 0.07 & 0.15 & \textbf{0.27} & \textbf{0.61} & \textbf{0.72} & \textbf{0.38} & 0.17 \\
14 & 157 & 1036 & 0.10 & 0.20 & 0.42 & 0.89 & 0.36 & 0.16 & 0.07 \\
15 & 331 & 1139 & 0.09 & 0.18 & 0.31 & 0.65 & 0.69 & 0.36 & 0.16 \\
16 & 623 & 1259 & 0.09 & 0.17 & 0.31 & 0.63 & 0.72 & \textbf{0.38} & 0.17 \\
17 & 235 & 566 & 0.10 & 0.25 & 0.58 & 0.86 & 0.51 & 0.24 & 0.10 \\
18 & 950 & 458 & 0.11 & 0.26 & 0.66 & 0.65 & 0.29 & 0.15 & 0.07 \\
19 & 1237 & 158 & 0.11 & 0.20 & 0.41 & 0.90 & 0.41 & 0.20 & 0.11 \\
20 & 850 & 1011 & 0.07 & 0.15 & 0.31 & 0.75 & 0.57 & 0.25 & 0.11 \\
\bottomrule
\end{tabular}}
\end{center}
\end{table}
We observed that in RGB space the change for all three metrics is approximately linear, which provides some advantage over using the same analysis in YUV space where the change for all three metrics is approximately exponential as shown in Figure \ref{fig:HI_KL_DB_YUV_safe}, Appendix \ref{appendix:supporting_materials}. Two cyan vertical lines where plotted at negative and positive 40, to represent a "safe" shift range where the autonomous system is expected to produce reliable predictions, as shown in Figures \ref{fig:dataShiftedSteeringPredictions-0m40m80m120} through \ref{fig:nvidia1_dataShiftedSteeringPredictions-p40p80p120}, Appendix \ref{appendix:supporting_materials}.

We then performed an experiment with partial results shown in Tables \ref{table:hi_rgb_space} through \ref{table:bhattacharyya_distance_yuv}. We chose fifty random image pairs from the Generated Track dataset (20 are displayed) where column ID1 is the frame number of the first random image, and ID2 is the frame number of the second random image.  

To the second random image ID2 with apply six shifts, ranging from negative to positive 120, and include the ID2 image with no shift, measuring the Histogram Intersection between both ID1 and ID2 for all shifts and making a record of the quantity, where values closer to 1, or most intersection, are the nearest, and closest to zero, or least intersection, represent the furthest distance between the pair of images. 

\begin{table}[ht!]
\caption{Relative Entropy in RGB space between random Generated Road frames}
\label{table:re_rgb_space}
\begin{center}
\resizebox{\columnwidth}{!}{%
\begin{tabular}{llllllllll}
\toprule
& ID1 & ID2 & -120 & -80 & -40 & 0 & 40 & 80 & 120 \\
\midrule
1 & 592 & 503 & 1.80 & 1.22 & 0.51 & 0.04 & 0.32 & 1.01 & 1.53 \\
2 & 863 & 825 & 1.82 & 0.98 & 0.34 & 0.07 & 0.83 & 1.58 & 1.97 \\
3 & 912 & 879 & 1.69 & 0.94 & 0.45 & 0.03 & 0.83 & 1.58 & 1.97 \\
4 & 1096 & 410 & 1.49 & 0.70 & \textbf{0.22} & \textbf{0.24} & \textbf{0.98} & 1.54 & 1.90 \\
5 & 519 & 772 & 1.71 & 0.97 & 0.32 & 0.07 & 0.67 & 1.38 & 1.81 \\
6 & 365 & 1091 & 1.82 & 1.23 & 0.64 & 0.16 & 0.25 & 0.78 & 1.43 \\
7 & 1082 & 1094 & 1.57 & 0.90 & 0.48 & 0.01 & 0.62 & 1.11 & 1.83 \\
8 & 811 & 1300 & 1.99 & 1.30 & 0.75 & 0.20 & 0.17 & 0.71 & 1.18 \\
9 & 146 & 1004 & 1.63 & 0.92 & 0.50 & 0.03 & 0.54 & 1.29 & 1.80 \\
10 & 1199 & 593 & 1.47 & \textbf{0.68} & 0.22 & 0.18 & 0.82 & 1.43 & 1.86 \\
11 & 797 & 1157 & 1.95 & 1.33 & \textbf{0.74} & \textbf{0.20} & \textbf{0.19} & 0.84 & 1.46 \\
12 & 350 & 173 & 1.88 & 1.31 & 0.71 & 0.17 & 0.17 & 0.72 & 1.37 \\
13 & 407 & 1260 & 1.93 & 1.33 & \textbf{0.74} & \textbf{0.20} & \textbf{0.16} & 0.68 & 1.32 \\
14 & 157 & 1036 & 1.64 & 0.90 & 0.46 & 0.04 & 0.81 & 1.63 & 2.00 \\
15 & 331 & 1139 & 1.83 & 1.25 & 0.68 & 0.16 & 0.25 & 0.80 & 1.41 \\
16 & 623 & 1259 & 1.85 & 1.28 & 0.69 & 0.17 & 0.17 & \textbf{0.65} & 1.26 \\
17 & 235 & 566 & 1.70 & 1.09 & 0.35 & 0.03 & 0.46 & 1.12 & 1.65 \\
18 & 950 & 458 & 1.63 & 0.82 & 0.22 & 0.18 & 1.03 & 1.60 & 1.97 \\
19 & 1237 & 158 & 1.63 & 0.91 & 0.48 & 0.02 & 0.67 & 1.15 & 1.81 \\
20 & 850 & 1011 & 1.90 & 1.16 & 0.64 & 0.08 & 0.32 & 1.13 & 1.72 \\
\bottomrule
\end{tabular}}
\end{center}
\end{table}

\begin{figure*}
\centering
\includegraphics[width=\textwidth]{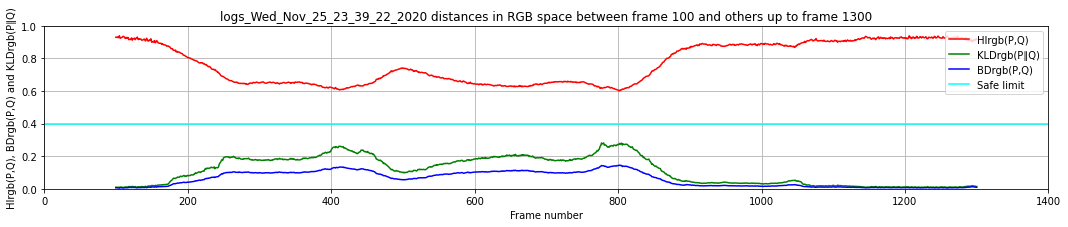}
\caption{Histogram Intersection, Relative Entropy and Bhattacharyya Distance in RGB space between frame 100 and others up to frame 1300, with a safe limit determined empirically, that is the safe distance threshold the Histogram Intersection should not fall below, or Relative Entropy and Bhattacharya distance should not go above}
\label{fig:Frames_100_1300_HI_KL_DB_safe}
\end{figure*}

Rows 11 and 13 are highlighted in Table \ref{table:hi_rgb_space} because the they represent the \textbf{largest} distance between ID1 and ID2 in the range of negative to positive 40, which is considered a "safe" shift. We highlight values the largest values in the negative and positive 80 columns, which represent the largest Histogram Intersection we would expect between any two pair of random images taken from e.g. a training dataset and an image acquired by the autonomous system in a real life situation. We could then claim that for any such pairing, we would accept the prediction of the network controlling the autonomous system as long as the Histogram overlap is greater than 0.40, ensuring that we are close to our safe limits, otherwise the autonomous system defers control to a human operator. The value 0.40 is approximately the y axis value for the x value of negative and positive 40, where the Histogram Intersection plot and the cyan "safe" vertical line intersect.

\begin{table}[ht!]
\caption{Bhattacharyya Distance in RGB space between random Generated Road frames}
\label{table:bhattacharyya_distance_rgb_space}
\begin{center}
\resizebox{\columnwidth}{!}{%
\begin{tabular}{llllllllll}
\toprule
& ID1 & ID2 & -120 & -80 & -40 & 0 & 40 & 80 & 120 \\
\midrule
1 & 592 & 503 & 1.59 & 0.85 & 0.32 & 0.03 & 0.15 & 0.54 & 1.06 \\
2 & 863 & 825 & 1.74 & 0.81 & 0.23 & 0.04 & 0.39 & 0.88 & 1.51 \\
3 & 912 & 879 & 1.66 & 0.85 & 0.32 & 0.01 & 0.40 & 0.83 & 1.43 \\
4 & 1096 & 410 & 1.19 & 0.50 & 0.13 & 0.12 & 0.49 & 0.91 & 1.48 \\
5 & 519 & 772 & 1.37 & 0.63 & 0.18 & 0.04 & 0.32 & 0.83 & 1.41 \\
6 & 365 & 1091 & 1.62 & 0.87 & 0.45 & 0.10 & 0.11 & 0.43 & 0.96 \\
7 & 1082 & 1094 & 1.51 & 0.75 & 0.35 & 0.01 & 0.30 & 0.64 & 1.23 \\
8 & 811 & 1300 & 2.09 & 1.12 & 0.59 & 0.14 & 0.09 & 0.41 & 0.80 \\
9 & 146 & 1004 & 1.55 & 0.75 & 0.35 & 0.02 & 0.29 & 0.69 & 1.21 \\
10 & 1199 & 593 & 1.13 & \textbf{0.48} & 0.13 & 0.10 & 0.42 & 0.82 & 1.36 \\
11 & 797 & 1157 & 2.17 & 1.20 & \textbf{0.59} & \textbf{0.14} & \textbf{0.09} & 0.45 & 0.95 \\
12 & 350 & 173 & 1.78 & 0.98 & 0.49 & 0.10 & 0.08 & 0.39 & 0.91 \\
13 & 407 & 1260 & 1.91 & 1.08 & \textbf{0.56} & \textbf{0.13} & \textbf{0.08} & 0.37 & 0.85 \\
14 & 157 & 1036 & 1.52 & 0.73 & 0.33 & 0.02 & 0.39 & 0.89 & 1.52 \\
15 & 331 & 1139 & 1.74 & 0.95 & 0.47 & 0.10 & 0.11 & 0.42 & 0.92 \\
16 & 623 & 1259 & 1.76 & 0.96 & 0.48 & 0.11 & 0.08 & \textbf{0.36} & 0.81 \\
17 & 235 & 566 & 1.43 & 0.72 & 0.22 & 0.02 & 0.24 & 0.65 & 1.22 \\
18 & 950 & 458 & 1.43 & 0.65 & 0.15 & 0.10 & 0.49 & 0.90 & 1.49 \\
19 & 1237 & 158 & 1.48 & 0.72 & 0.33 & 0.01 & 0.33 & 0.67 & 1.24 \\
20 & 850 & 1011 & 2.00 & 1.04 & 0.50 & 0.05 & 0.15 & 0.58 & 1.13 \\
\bottomrule
\end{tabular}}
\end{center}
\end{table}

We repeated the experiment for Relative Entropy, highlighting rows 4, 11 and 13, which represent the \textbf{largest} Relative Entropy values between ID1 and ID2, where larger is further (greater distribution shift) and smaller is nearer (smaller distribution shift). We highlight the smallest values in columns negative and positive 80 shift, and claim that as long as the Relative Entropy between a random image from the training dataset and one acquired from the autonomous system camera, is less than 0.60, we will accept the prediction of the network in control of the autonomous system, else the prediction is considered to be unsafe and control is delegated to a human operator.

We repeated the experiment in RGB space and computed the Bhattacharyya Distance, with results shown in table \ref{table:bhattacharyya_distance_rgb_space}. We use the same images for all distance metrics, i.e. row 1 will always be a comparison between frames 592 and 503 for all six, three RGB and three YUV tables, for the Histogram Intersection, Relative Entropy and Bhattacharyya Distance metrics, that is the 50 image pairs were chosen randomly only once, and the same list of pairs is used thereafter. Again we highlight the rows with the largest distances which are coincident for all three tables described so far, rows 11 and 13. Again we look at the smallest values in columns negative and positive 80 that a Bhattacharyya Distance of 0.30 or less, between a randomly chosen image from the training dataset, and an image acquired by the autonomous system's camera, is safe to use, and the network prediction controlling the autonomous system will be accepted, otherwise the prediction will be rejected, and control will be returned to a human operator.

We concluded that in the RGB space, any of the three distance metrics would be adequate, and the least computationally intensive would be preferred, which is the \textbf{Histogram Intersection}.

 \section{Conclusion}

A formulation could be given by defining a range of maximum and minimum distribution shift using pixels intensity value averages as a proxy, over the entire training dataset, and the allowed data distribution shift. The output being a boolean value where true means the image and resulting prediction are safe to use, false otherwise.

\begin{equation}
    P_{safe} = 
    \begin{cases}
    true , & \text{if $|D_Brgb(P,Q)| < |D_Brgb(P,Q_{sr})|$}.\\
    true, & \text{if $|D_\text{KLrgb}(P \parallel Q)| < |D_\text{KLrgb}(P \parallel Q_{sr})|$}.\\
    true, & \text{if $|H_{irgb}(P, Q)| > |H_{irgb}(P, Q_{sr})|$}. \\
    false ,  & \text{otherwise}.
    \end{cases}
\end{equation}
Where $P_{safe}$ is a boolean value determining if it is safe to make a prediction from input $Q$ relative to $Q_{sr}$, and $Q_{sr}$ is the safe range between negative and positive $sr$, and the comparisons are performed in RGB space.
Given the computational advantages, and strong linearity
$H_{irgb}$ is the preferred distance metric.

{\small
\bibliographystyle{ieee_fullname}
\bibliography{main}
}

\clearpage
\appendix

\section{Supporting Materials}
\label{appendix:supporting_materials}


\begin{figure*}[!ht]
\centering
\includegraphics[width=\textwidth]{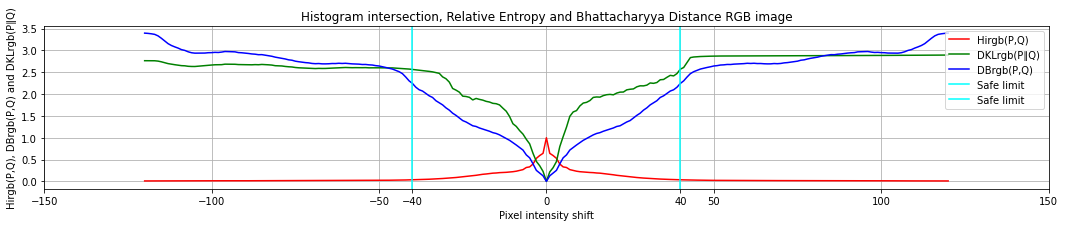}
\caption{Histogram Intersection, Relative Entropy and Bhattacharyya Distance safe limits for YUV image pixel intensity value shifts}
\label{fig:HI_KL_DB_YUV_safe}
\end{figure*}

\begin{table*}[ht!]
\caption{Bhattacharyya Distance in YUV space between random Generated Road frames}
\label{table:bhattacharyya_distance_yuv}
\begin{center}
\begin{tabular}{llllllllll}
\toprule
& ID1 & ID2 & -120 & -80 & -40 & 0 & 40 & 80 & 120 \\
\midrule
1 & 592 & 503 & 3.51 & 2.14 & 1.62 & 0.07 & 1.41 & \textbf{1.97} & 3.33 \\
2 & 863 & 825 & 3.66 & 2.74 & 1.37 & 0.10 & 2.00 & 2.76 & 4.39 \\
3 & 912 & 879 & 3.02 & 2.35 & 1.45 & 0.08 & 2.39 & 2.72 & 3.83 \\
4 & 1096 & 410 & 3.20 & 2.41 & 1.31 & 0.20 & 2.23 & 2.95 & 4.46 \\
5 & 519 & 772 & 2.90 & \textbf{1.84} & 1.35 & 0.13 & 1.63 & 2.44 & 3.59 \\
6 & 365 & 1091 & 3.56 & 2.74 & 1.73 & 0.16 & 1.37 & 2.34 & 3.29 \\
7 & 1082 & 1094 & 3.44 & 2.85 & 2.32 & 0.03 & 2.33 & 2.80 & 3.45 \\
8 & 811 & 1300 & 4.95 & 2.91 & 2.35 & 0.24 & 1.23 & 2.11 & 3.27 \\
9 & 146 & 1004 & 3.24 & 2.68 & 1.85 & 0.05 & 1.82 & 2.57 & 3.07 \\
10 & 1199 & 593 & 2.87 & 2.07 & 1.25 & 0.18 & 1.67 & 2.50 & 3.43 \\
11 & 797 & 1157 & 4.64 & 2.90 & \textbf{2.22} & \textbf{0.24} & \textbf{1.20} & 2.35 & 3.67 \\
12 & 350 & 173 & 4.12 & 3.16 & 1.85 & 0.20 & 1.38 & 2.29 & 3.32 \\
13 & 407 & 1260 & 4.71 & 3.01 & 2.37 & 0.18 & \textbf{1.20} & 2.01 & 2.87 \\
14 & 157 & 1036 & 3.44 & 2.70 & 2.38 & 0.06 & 2.29 & 2.94 & 3.89 \\
15 & 331 & 1139 & 3.64 & 2.66 & 1.81 & 0.15 & 1.32 & 2.25 & 3.21 \\
16 & 623 & 1259 & 3.73 & 2.78 & 1.72 & 0.18 & 1.28 & 2.16 & 2.98 \\
17 & 235 & 566 & 3.54 & 2.88 & 1.55 & 0.11 & 1.44 & 2.36 & 3.87 \\
18 & 950 & 458 & 2.88 & 2.06 & 1.20 & 0.17 & 2.24 & 2.81 & 4.35 \\
19 & 1237 & 158 & 4.72 & 2.95 & 1.93 & 0.06 & 2.28 & 2.75 & 3.47 \\
20 & 850 & 1011 & 4.45 & 2.78 & 2.30 & 0.13 & 1.50 & 2.02 & 2.71 \\
\bottomrule
\end{tabular}
\end{center}
\end{table*}

\begin{figure*}[!ht]
\centering
\includegraphics[width=\textwidth]{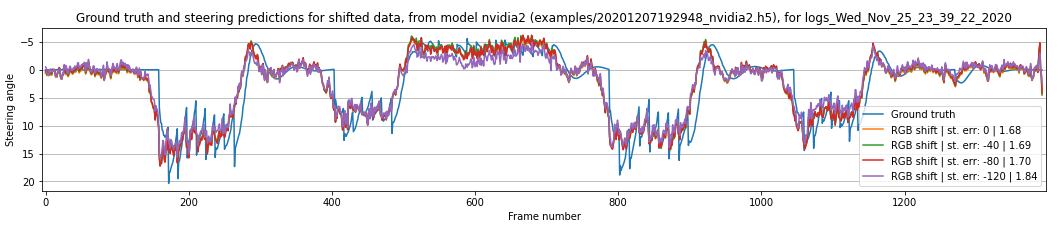}
\caption{Plots of ground truth (SDSandbox PID steering output) and nvidia2 network predictions for images with pixel value intensity shifts of 0, -40, -80 and -120}
\label{fig:dataShiftedSteeringPredictions-0m40m80m120}
\end{figure*}


\begin{figure*}[!ht]
\centering
\includegraphics[width=\textwidth]{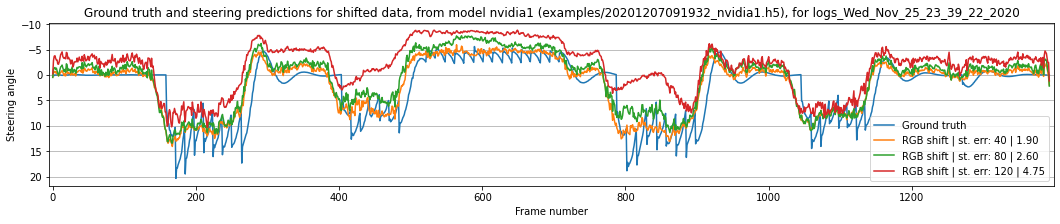}
\caption{Plots of ground truth (SDSandbox PID steering output) and nvidia1 network predictions for images with pixel value intensity shifts of 40, 80 and 120}
\label{fig:nvidia1_dataShiftedSteeringPredictions-p40p80p120} 
\end{figure*}

\end{document}